# Modeling and Simulation of the Quadrtor Delivery System

A.M. El-Edkawy and M.A. El-Dosuky

Computer Science Department, Faculty of Computers and Information, Mansoura University, Egypt

amr.eledkawy@mans.edu.eg, mouh_sal_010@mans.edu.eg

**Abstract:** *Unmanned Aerial Vehicle (UAV) quadrotor is a UAV with four rotors. The quadrotor control is a difficult task because the four-wheel system is inefficient. The purpose of this paper is to provide specifications and implement a separate quad system that integrates data from the camera and IMU (Inertial Measurement Unit) to allow high-resolution optical transmission. In this paper, a separate quad system was proposed. The system specification is specified. The system application introduced the SolidWorks design of the system, the mathematical modeling of the pickup and delivery (PDP) problem. The simulation results using MATLB for model implementation and the simultaneous localization and mapping (SLAM) were presented using the Extended Kalman (EKF-SLAM) filter algorithm used to make the system independent. Experimental results were introduced to use the candy edge detection algorithm.*

## 1. Introduction

Unmanned Aerial Vehicle (UAV) quadrotor is a UAV with four rotors. Enabling quadrotor mobility independently in external environments is very difficult. It is difficult to control quadrotor, because it is considered an inefficient system [2, 3] as it contains six degrees of freedom (DOF) but only four triggers. The integration of inertial measurements and features in images taken from the camera is known as inertial inertial navigation [4, 5].

The purpose of this paper is to provide specifications and implement a separate quadrotor system that integrates data from the camera and IMU (Inertial Measurement Unit) to allow high-resolution optical transmission. In this paper, a separate quadrotor system was proposed. The system specification is specified. The system application introduced the SolidWorks design of the system, the mathematical modeling of the pickup and delivery problem (PDP) [6, 7]. The simulation results using MATLAB for the implementation of the simultaneous localization and mapping (SLAM) were presented using the Extended Kalman (EKF-SLAM) filter algorithm used to make the system independent. Experimental results were introduced to use the edge detection algorithm of the enclosure [9, 10].

Later in the paper, the second section explains the background and previous work on the design requirements of the quadcopter delivery, modeling and control, how to take advantage of the computer's vision algorithm and how to make the system independent. Section 3 outlines the proposed system by providing a specification and drawing a general framework for explaining it, then presenting the SolidWorks model and mathematical modeling for the PDP. Section 4 illustrates the simulation and results developed using MATLAB. Section 5 summarizes the paper and provides future guidance based on the proposed system.

## 2. Related Work

Prior to this, in an ideal world, a quadcopter will be equipped for independent flight through corridors, rooms and stairs in any building. In the building, a quadcopter should be suitable for transferring data about the environment remotely to the first remote interaction group. The following are the basic requirements for UAV [11]:

1. Battery life is at least 10 minutes.
2. Move in 40-50% of the throttle limit (1.5: 1-2: 1 to the weight ratio).
3. Sensors to estimate the relative intervals of distant objects, length and distance measurement.
4. Independent taking off, hovering, traversing and falling.
5. Wireless availability.
6. Maximum width below 30 "(normal width of the gate).
7. Total mass under 1.5 kg, for possible load of 0.5kg-1kg
8. Protect the fans with the ultimate goal that the quadcopter can survive after falling 10 feet.
9. The flight height is a little about 5-7ft, the maximum height is at least 10 feet.

There are a few other options to create and develop crafts that are worth considering, but everyone has their cons. Most prominently, commercial drones tend to have some basic categories of vulnerabilities. They have a tendency to be intended to transport any additional load other than themselves (eg customer level Draganflyer, AR.Drone, Parrot AR.Drone, Silverlit X-UFO, etc.), while those that can transfer extra are too expensive (Microdrone MD4-200 , Modern Draganflyer, AscTec), or it is so huge that it can not work out in the case of our own target use.

Open implementation processes, for example, can be Aeroquad and MikroKopter, through extensive modification of the design, near our desirable standards but will spare a great effort of the designer and will reduce the value of the platform and its long-term adaptability and age [11].

### 2.1 Modeling and Control of Quadrotor

Quadrotor's main idea was to have a quadcopter that could handle larger limits and could move in areas that were hard to reach. It has four fans (right, left, front and back). The speed of the motors is changed to control the quadrocopter in each of the three axes. As shown in Figure 1, the right and left motors rotate counterclockwise while the front and rear fans rotate clockwise. This adjusts the total torque and cancels aerodynamics and torque in fixed flights [12].

Yaw: Changing rpms to reverse-engine rotor sets, the quadrocopter will bend due to the moment it was created.

Roll: Increasing and decreasing RPM for left and right engines makes the copter spin forward or backward.

Pitch: Like a roll but using front and rear engines instead of left and right.

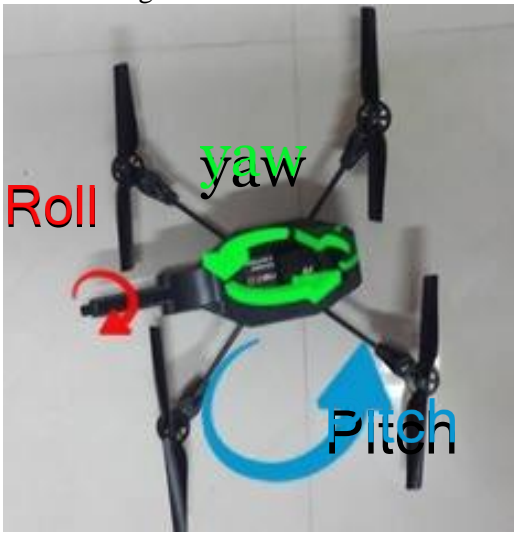

Fig 1: Roll, Yaw and Pitch

There are several papers that discussed the Quadrotor Modeling [13-17]. In [17] the final result reached for the equations of motion under the structure of the quadrilateral are as follows:

$$\dot{\omega} = \begin{bmatrix} \tau\varphi I_{xx}^{-1} \\ \tau\theta I_{yy}^{-1} \\ \tau\psi I_{zz}^{-1} \end{bmatrix} - \begin{bmatrix} \frac{I_{yy}-I_{zz}}{I_{xx}}\omega_y\omega_z \\ \frac{I_{zz}-I_{xx}}{I_{yy}}\omega_x\omega_z \\ \frac{I_{xx}-I_{yy}}{I_{zz}}\omega_x\omega_y \end{bmatrix} \quad (1)$$

where τ is the motor torque, $I$ is the input current, $\varphi, \theta$ and $\psi$ are the roll, pitch, yaw in the body frame respectively, $x, y$ and $z$ are determining the position of the quadrotor, $\omega$ is the angular velocity vector in the body frame.

In order to simulate the real world, we must perform quadratic control, showing the Proportional-Integral-Derivative (PID) control that controls position, position and quadrature height. The following figure shows the PID controller [14].

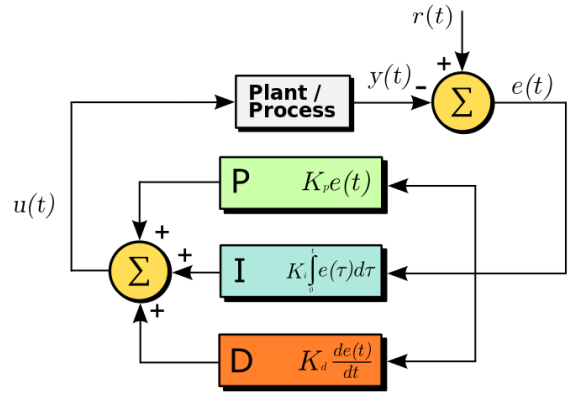

Fig 2: PID Control

### 2.2 Canny Edge Detection Vision

Canny edge detection algorithm is a popular multi step edge detection algorithm starting with input image $f(u, v)$ of size $u \times v$ pixels. It has the following steps [9, 10].

1. Apply Gaussian filter $G_a$ to smooth the image $f(u, v)$ in order to remove the noise

$$gr(u, v) = G_a(u, v) * f(u, v) \quad (2)$$

where

$$G_a = \frac{1}{\sqrt{2x\sigma^2}} \exp\left(-\frac{u^2 + v^2}{2\sigma^3}\right) \quad (3)$$

2. Find the intensity gradients $M(u, v)$ of the smoothed image $gr(u, v)$

$$M(u, v) = \sqrt{gr_u^2(u, v) + gr_v^2(u, v)} \quad (4)$$

and the gradient direction

$$\theta(u, v) = \tan^{-1}\{gr_v(u, v)/gr_u(u, v)\} \quad (5)$$

3. Threshold

$$M_T(u, v) = \begin{cases} M(u, v) & \text{if } M(u, v) > T \\ 0 & \text{otherwise} \end{cases} \quad (6)$$

4. Apply the maximum suppression to eliminate the counterfeit response to edge detection
5. Apply double threshold to select possible edges
6. Apply hysteresis: Finish the edge detection by eliminating the weak edges.

### 2.3 Autonomous Robot

Figure 3 shows the cluster block diagram in most independent robots.

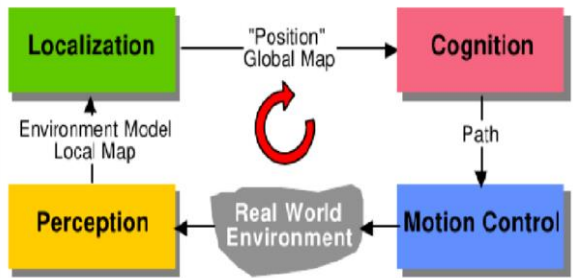

Fig 3: Autonomous robots control [18]

To be independent, the quadrotor must solve the problem of SLAM [8]. So they have to locate it within a different map you create from the specific environment on synthesizing the data [19] from many of the data sources disturbing [1]. Applications utilize sonar [20] or optical navigation [21]. The normal user filter is the Kalman filter [22]. The RaoBlackwellized particle filter (FastSLAM) and EKF-SLAM were illustrated in [8]. In random mapping [23], the data association tends to use a traditional procedure to follow the problems known as the nearest neighbor (NN) [24]. The development of the Joint Compatibility Branch-and-Bound data association (JCBB) [25] is supported by the way the NN algorithm is used to link data that is exceptionally sensitive to sensor error, increasing the likelihood of unattended map features [25].

## 3. Proposed System

### 3.1 Specification and Setup

PEAS (performance meter, environment, actuators, sensors) [26] is used to determine proxy specifications. In the proposed system, the factor is quadruple. Table 1 shows the PEAS specification for a quadcopter.

Table 1: PEAS of the quadcopter

| Performance Measure | Reaching destination quickly and correctly |
|---|---|
| Future Performance Measure | Collaboration among quadrotors as a swarm |
| Environment | Roads, Buildings and objects |
| Actuators | 4 rotors |
| Sensors | Camera and Inertia Measurement Unit (IMU) |

ODESA (observability, deterministic, episodic, static, agents) [26] is used to describe the characteristics of the environment. Table 2 specifies the properties of the quadrotor environment.

Table 2: ODESA of the quadrotor environment

| Observable | Partial Observable |
|---|---|
| Deterministic | Stochastic |
| Episodic | Sequential |
| Static | Dynamic |
| Agents | Multi-agent |

Figure 4 shows the system block diagram that shows a comprehensive system overview.

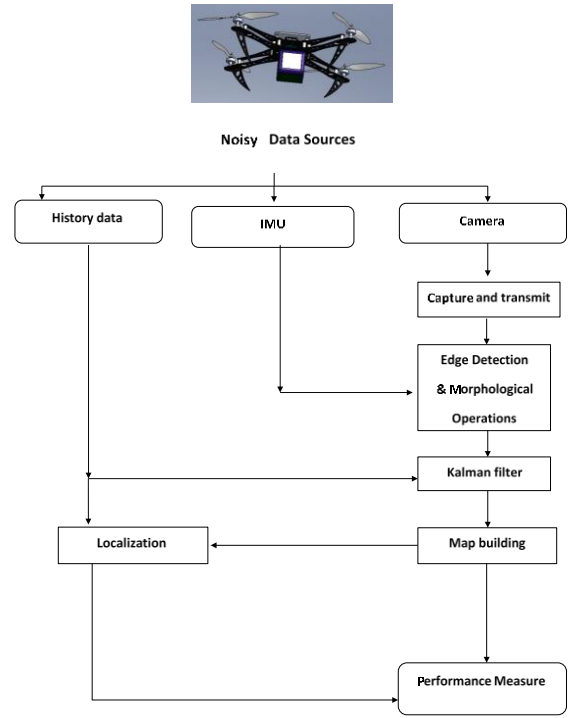

Figure 4: Proposed System

### 3.2 Design and Model

Figure 5 illustrates our three-dimensional model of the Quad in SolidWorks. This form contains the following components.

Table 3: 3D model components

| 2 Bodies |
|---|
| Arduino-UNO |
| Battery 4Ah 5s Lipo Nano-Tech Jr. |
| Spectrum receiver BR 60000 Kevin Barker |
| 4 Propellers |
| 4 Motors 1300Kv |
| 4 Above Paws |
| 4 Propeller Screws |
| 4 Down Paws |
| 4 Paws |
| 4 Block Paws |
| 12 Separators |
| Camera |
| IMU |

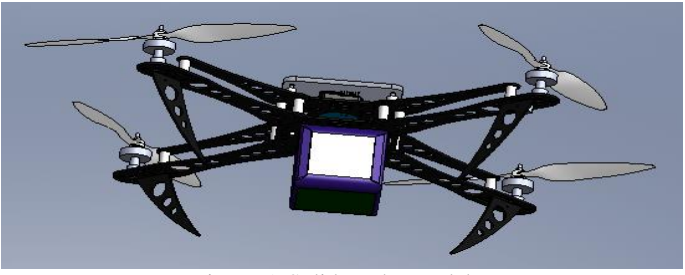
Figure 5: SolidWorks Model

The quadrotor model has been improved with the problem of finding a quadrotor path that represents a variable for pockup and delivery (PDP) problem. It is expressed in [6] as a linear program with many limitations in the delivery system.

### 3.3 PDP

The problem of finding a quadrotor path is the alternative to PDP. They are expressed here as a linear program. We will use the following nomenclature:

- $V$ is a collection of all ground vehicles (base stations of a helicopter).
- $H$ is a collection of all quadcopters.
- $S$ is a collection of all elements in the request system.
- $R$ is a collection of requests that must be served.
- $N = H \cup S \cup V$ is the collection of all important sites
- $E = N^2$ is a collectionof directed edges in the graph on $N$

For $e \in E$, we will denote the distance between the endpoints of the edge by $|e|$. Sometimes, we will refer to an edge by its ordered pair. The variables of the linear program are:

- $\{x_e^h\}_{e \in E, h \in H}$ Binary variables with value 1 if quadcopter h traverses edge e and 0 otherwise
- $\{q_e^s\}_{e \in E, s \in S}$ Quantity of item s carried along edge e (integer)
- $\{z_e^h\}_{e \in E}^{h \in H}$ Quantity representing the state of the battery of quadcopter h after traversing edge e
- $\{t_e^h\}_{e \in E, h \in H}$ Abstract quantity roughly quantifying the number of legs remaining in the trip for quadcopter h after traversing edge e.
- $T$ Time required for all quadcopters to complete their routes

The objective function is then easily expressed as, $f = T$

**Restrictions**

There are a large number of restrictions, so we will provide associated restrictions:

- $\forall h \in H, e \in E, \quad 0 \leq x_e^h \leq 1$
- $\forall s \in S, e \in E, \quad 0 \leq q_e^s$
- $\forall h \in H, \forall e \in E \quad 0 \leq z_e^h$
- $\forall h \in H, e \in E, \quad 0 \leq t_e^h$
- $\forall h \in H, v \in N, \quad x_{(v,v)} \leq 0$ (Prevent self-loops)
- $\forall h \in H, v' \in N, \quad x_{(v',v)}^h \leq 0$, where quadcopter $h$ starts at $v$. (No one can fly to the starting location of a quadcopter)
- $\forall h \in H, v' \in N, \quad z_{(v,v')} \leq c - \frac{|(v,v')|}{maxRange}$, where quadcopter $h$ starts at $v$ with initial charge, $c$. (Remaining battery after the first leg is bounded by the initial charge less the battery consumed on the first leg)
- $\forall v \in V, v' \in N, \quad z_{(v,v')}^h \leq 1 - \frac{|(v,v')|}{maxRange(h)}$ (The remaining battery after leaving the station is fully charged, minus the battery consumed in leaving the station)
- $\forall h, h' \in H, v' \in N, x_{v,v'}^{h'} \leq 0$, where quadcopter $h$ starts at $v$ and $h \neq h'$. (No quadrotor may leave the start site as another quadcopter)

The remaining equality and inequality restrictions are divided into a number of species.

- Type 0 constraints. Each quadcopter can only leave the starting position in one direction.

$$\forall h \in H, \sum_{e=(v,v') \in E} x_e^h \leq 1,$$
where quadcopter $h$ starts at $v$.

- Type 1 restrictions. The amount of each element moved away from each site starting a quadcopter is equal to the amount of that element that a given quadcopter currently carries.

$$\forall h \in H, s \in S, v' \in N,$$
$$q_{(v,v')}^s = x_{(v,v')}^h q_{init}(h),$$ where quadcopter $h$ starts at $v$

- Type 2 restrictions. There is no air route twice.

$$\forall e \in E, \sum_{h \in H} x_e^h \leq 1$$

- Type 3 restrictions. Each request is served.

$$\forall r \in R, s \in S, \sum_{e=(v',v) \in E} q_e^s - \sum_{e=(v,v') \in E} q_e^s = \# \text{ of items of type } s \text{ in request } r$$

Here we use the term that delivery is a positive number of elements and that the pick-up is negative.

Also, $v$ corresponds to the location of request $r$.

- Type 4 restrictions. Each request is visited once

$$\forall r \in R, \sum_{h \in H} \sum_{e=(v',v) \in E} x_e^h = \sum_{h \in H} \sum_{e=(v,v') \in E} x_e^h = 1, \text{ where } v \text{ is the location of request } r$$

- Type 5 restrictions. Shipments must respect the absorptive capacity of the quadcopter.

$$\forall e \in E, \sum_{s \in S} q_e^s weight(s) - \sum_{h \in H} x_e^h capacity(h) \leq 0$$

- Type 6 restrictions. The quadcopter travels to a request if and only if it is moved away.

$$\forall r \in R, h \in H, \sum_{e=(v',v) \in E} x_e^h - \sum_{e=(v,v') \in E} x_e^h$$

- Type 7 restrictions. A quadcopter reaches each station at least several times as you leave.

$$\forall h \in H, v \in V, \sum_{e=(v,v') \in E} x_e^h - \sum_{e=(v',v) \in E} x_e^h \leq 0$$

- Type 8 restrictions. The battery is only used on the legs being moved.

$$\forall h \in H, \forall e \in E, \quad z_e^h \leq x_e^h maxcharge(h)$$

- Type 9 restrictions. The amount of battery remaining after leaving a request is equal to the amount of battery when this demand is reached less amount of battery consumed in leaving the demand

$$\forall h \in H, \forall r \in R, \sum_{e=(v',v) \in E} z_e^h - \sum z_e^h = \sum x_e^h |e|$$

where $v$ is the site of request $r$.

- Type 10 restrictions. The number of remaining legs in the path is logical for the edges in the path only.

$$\forall h \in H, e \in E, \quad t_e^h \leq x_e^h(2|R| + |V|)$$

Here, $2|R| + |V|$ was chosen as an upper bound for the number of legs in a minimal route, effectively leaving $t_e^h$ unbounded wherever $x_e^h$ is positive.

- Type 11 restrictions. The number of legs remaining after leaving the node (other than the prefix node) equals the number of remaining legs when reaching that node minus the number of departures from the node.

$$\forall v \in V \cup R, h \in H, \sum_{e=(v',v) \in E} t_e^h - \sum t_e^h = \sum x_e^h$$

where $v$ is the location of request $r$. Although this is not clear, it is true even if a quadrtor visits a node several times. The first and last departures have meaningful values even though any intermediate circuits may have interchangeable values.

- Type restrictions 12. Total time taken is at least as long as each quadcopter takes. (This is how we have restricted the maximum ())

$$\forall h \in H, \sum_{e \in E} x_e^h |e| \leq T$$

For the actual implementation, variables are arranged into $N \times N$ blocks, corresponding to the edge index. Within each block, the nodes are arranged in the order, request locations, quadcopter start locations, then ground vehicle locations. The variables are placed in column-major order within each block. The blocks are then arranged in order with x values first, followed by q's, then z's, then t's.

For example, suppose we are dealing with a single helicopter, one request, one item, and one floor car. The demand will then be on site 1, the helicopter on site 2, and the ground vehicle at location 3. The variables in the linear program will be arranged in order,

$x_{11}^1, x_{21}^1, x_{31}^1, x_{12}^1, x_{22}^1, x_{32}^1, x_{31}^1, x_{32}^1, x_{33}^1,$
$q_{11}^1, q_{21}^1, q_{31}^1, q_{12}^1, q_{22}^1, q_{32}^1, q_{31}^1, q_{32}^1, q_{33}^1,$
$z_{11}, z_{21}, z_{31}, z_{12}, z_{22}, z_{32}, z_{31}, z_{32}, z_{33},$
$t_{11}^1, t_{21}^1, t_{31}^1, t_{12}^1, t_{22}^1, t_{32}^1, t_{31}^1, t_{32}^1, t_{33}^1,$ T

We can also make a number of initial cuts. These are the inequalities that arise from integrative constraints, but not from linear constraints. Including these reductions are not necessary, but they restrict linear relaxation, which can reduce the time of solution.

- Additional limits. The legs that depart from the helicopter launch site can be restricted on charges. If the leg end is a delivery and the quadcopter shipments do not include the goods to be delivered, then that leg can not be taken. Similarly, if the end of the leg is a pickup and there is not enough space in the capacity of a helicopter, the leg can not be taken.

- Type 13 restrictions. All quadcopters must move. If a helicopter is already in a vehicle, we want it to wait, but this can be achieved by moving a distance of 0 to the vehicle.

$$\forall h \in H, \sum_{v' \in N} x^h_{start(h),v'} \geq 1$$

## 4. Results

MATLAB is used to simulate the delivery quadrotor.

### 4.1 Model Simulation

The quadrotor model simulation shown in figure 6.

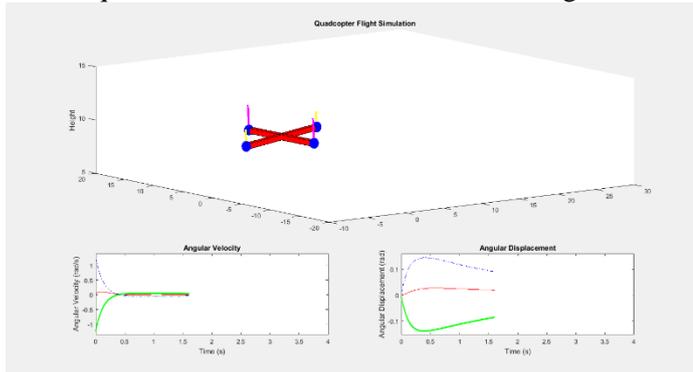

Fig 6: Model Simulation

### 4.2 Simulink Models

The followings are snapshots of the Simulink models

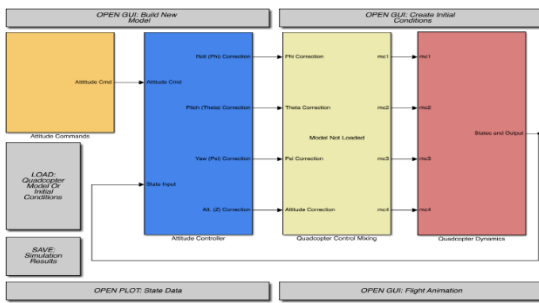

Fig 7: Attitude control Simulink model

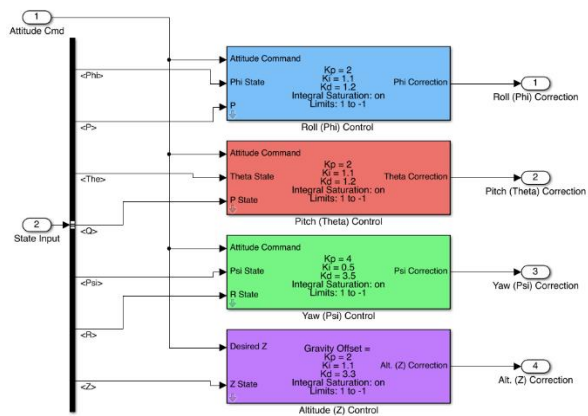

Fig 8: Attitude controller

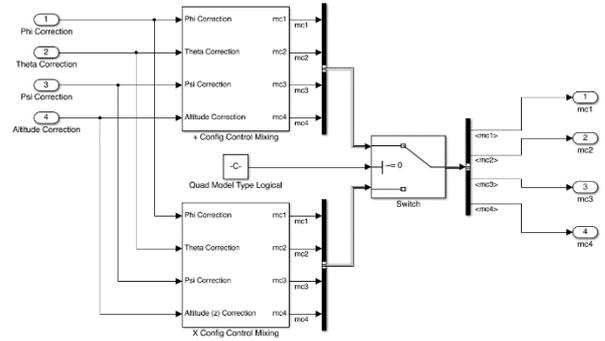

Figure 9: Quadcopter control mixing overview

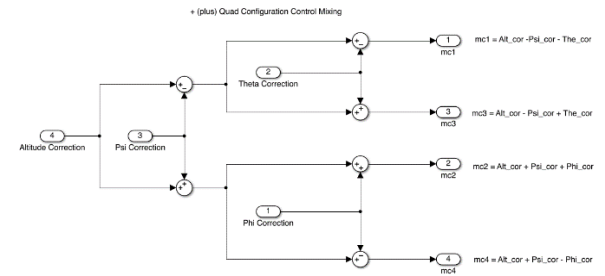

Fig 10: Plus configuration control mixing

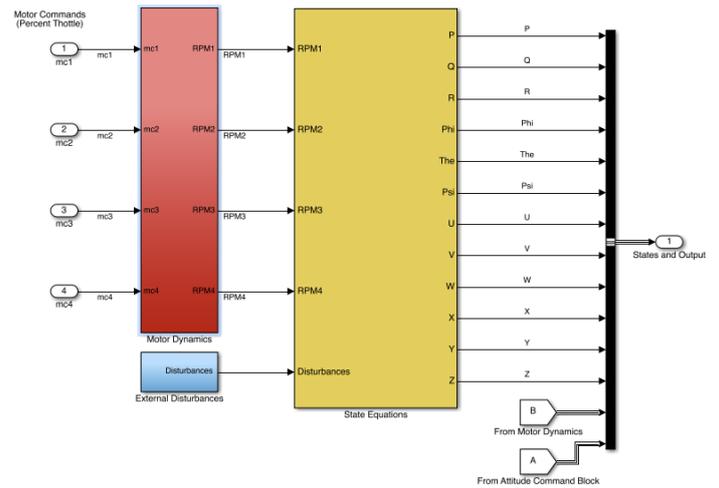

Fig 11: Quadcopter dynamic block

### 4.3 Attitude and Position Control (AC, PC) Results

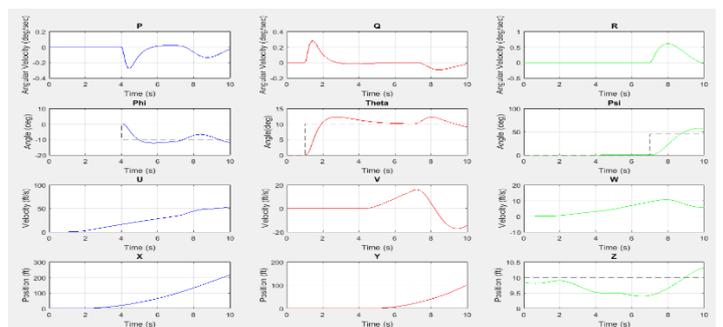

Fig 12: Position, velocity, angle and angular velocity results for AC

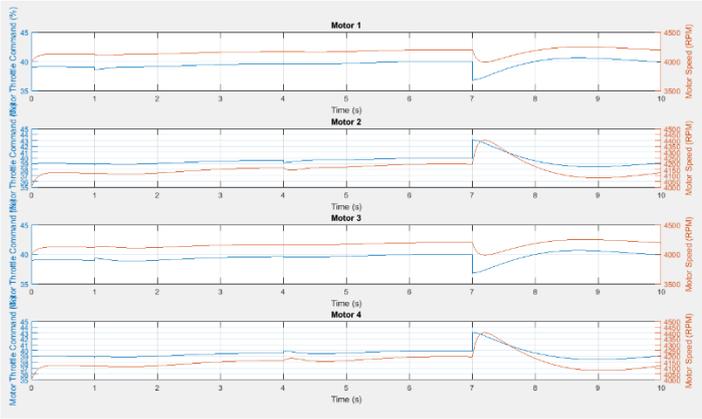
Fig 13: Motor results for AC

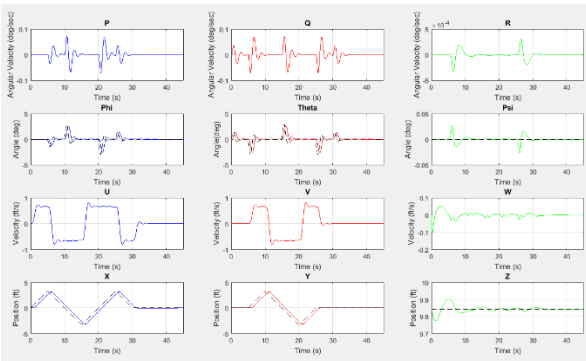
Fig 14: Position, velocity, angle and angular velocity results for PC

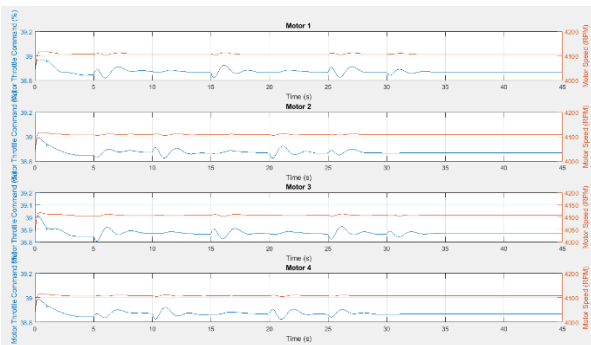
Fig 15: Motor results for PC

### 4.4 Computer Vision

The quadrocopter system records and loads images and navigation data and videos instantly to be handled first by getting the edge using edge detection [9, 10] as shown in Figure 16. Then through the data integration application, the system can navigate well by maintaining The faraway is exactly on its way around the walls.

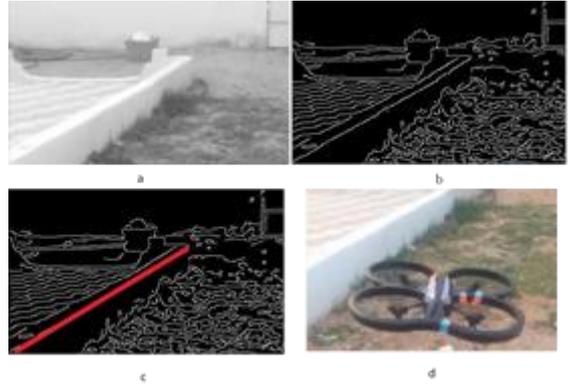
Fig 16: Canny edge detection

The quadrotor updates its position on a local map as shown in the following section.

### 4.5 SLAM

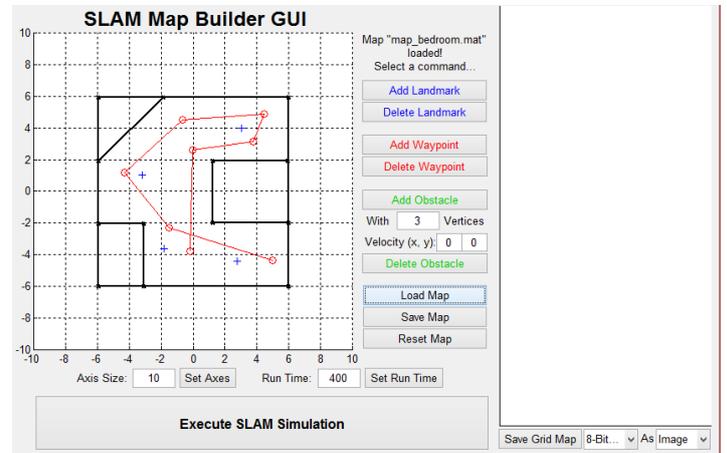
Fig 17: SLAM Map Builder

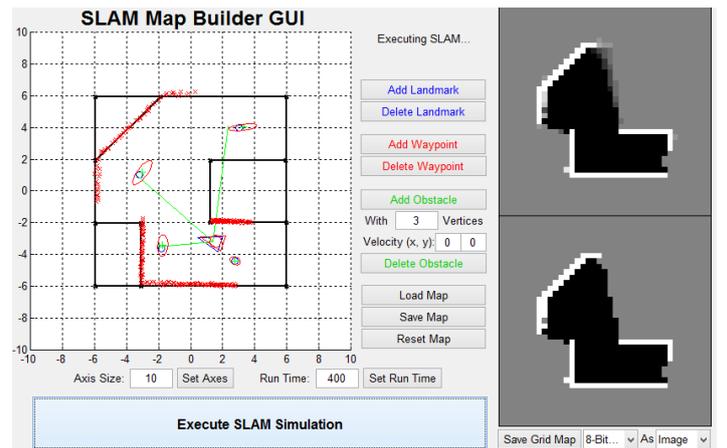
Fig 18: SLAM in progress

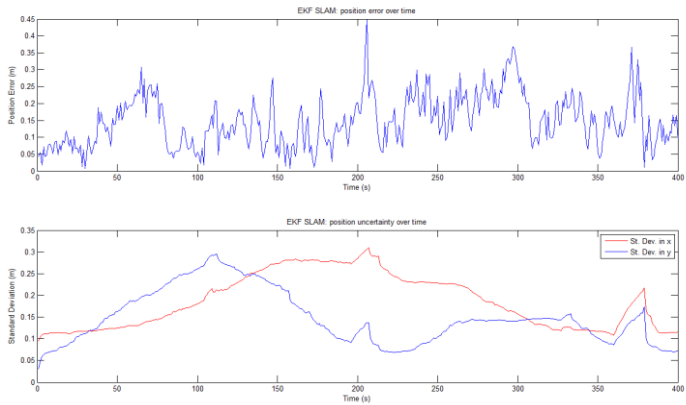

Fig 19: Position error and uncertainty for SLAM

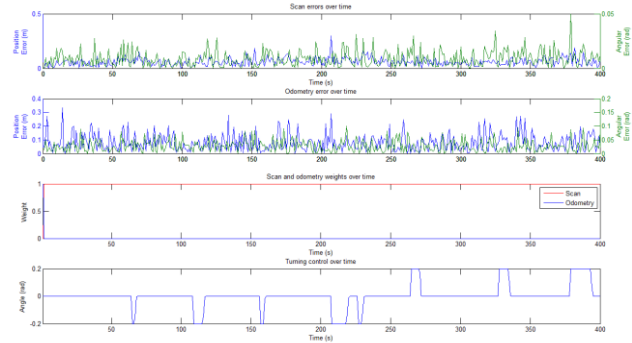

Fig 20: SLAM results

## 5. Conclusion and Future Work

The purpose of this paper is to provide specifications and implement a separate quad-band system that integrates data from the camera and IMU to allow high-resolution optical transmission.

In this paper, a separate quad-band system was proposed. The system specification is specified. The system application introduced the SolidWorks design of the system, the mathematical modeling of the PDP problem.

The simulation results using MATLAB for model implementation and SLAM were presented using the EKF-SLAM filter algorithm used to make the system independent. Experimental results were introduced to use the candy edge detection algorithm.

Future work involves the use of solar energy for charging and the establishment of co-ordination between the quadrotors.

## References

[1] Lozano, Rogelio, ed. Unmanned aerial vehicles: Embedded control. John Wiley & Sons, 2013.

[2] Hou, Hongning, et al. "A simple controller of minisize quad-rotor vehicle." Mechatronics and Automation (ICMA), 2010 International Conference on. IEEE, 2010.

[3] Kim, Jinhyun, Min-Sung Kang, and Sangdeok Park. "Accurate modeling and robust hovering control for a Quad–rotor VTOL aircraft." Journal of Intelligent and Robotic Systems 57.1-4 (2010): 9.

[4] Mourikis, Anastasios I., et al. "Vision-aided inertial navigation for spacecraft entry, descent, and landing." IEEE Transactions on Robotics 25.2 (2009): 264-280.

[5] Kottas, Dimitrios G., et al. "On the consistency of vision-aided inertial navigation." Experimental Robotics. Springer International Publishing, 2013.

[6] Mosterman, Pieter J., et al. "A heterogeneous fleet of vehicles for automated humanitarian missions." Computing in Science & Engineering 16.3 (2014): 90-95.

[7] Hartman, M. M. S. M. D., K. Landis, and J. K. D. B. C. Chang. "Quadcopter dynamic modeling and simulation." freeware project presented at "2014 MATLAB and Simulink Student Design Challenge". 2015.

[8] Durrant-Whyte, Hugh, and Tim Bailey. "Simultaneous localization and mapping: part I." IEEE robotics & automation magazine 13.2 (2006): 99-110.

[9] Green, Bill. "Canny Edge Detection Tutorial.", (2002).

[10] Bailey, Mark Willis. Unmanned aerial vehicle path planning and image processing for orthoimagery and digital surface model generation. Diss. Vanderbilt University, 2012.

[11] D'Angelo, R., and R. Levin. "Design of an Autonomous Quad-rotor UAV for Urban Search and Rescue." Worcester Polytechnic Institute (2011).

[12] Bouabdallah, Samir, Andre Noth, and Roland Siegwart. "PID vs LQ control techniques applied to an indoor micro quadrotor." Intelligent Robots and Systems, 2004.(IROS 2004). Proceedings. 2004 IEEE/RSJ International Conference on. Vol. 3. IEEE, (2004).

[13] Bresciani, Tammaso. "Modelling, identification and control of a quadrotor helicopter." MSc Theses, Department of Automatic Control, Lund University (2008).

[14] Bouabdallah, Samir, Pierpaolo Murrieri, and Roland Siegwart. "Design and control of an indoor micro quadrotor." Robotics and Automation, 2004. Proceedings. ICRA'04. 2004 IEEE International Conference on. Vol. 5. IEEE, (2004).

[15] Johnson, Eric N., and Michael A. Turbe. "Modeling, control, and flight testing of a small ducted-fan aircraft." Journal of Guidance Control and Dynamics 29.4 (2006): 769-779.

[16] Hoffmann, Gabriel M., et al. "Quadrotor helicopter flight dynamics and control: Theory and experiment." Proc. of the AIAA Guidance, Navigation, and Control Conference. Vol. 2. (2007).

[17] Gibiansky, Andrew. "Quadcopter dynamics, simulation, and control." Andrew Gibiansky:: Math→[Code] 21 (2012).

[18] Siegwart, Roland, Illah Reza Nourbakhsh, and Davide Scaramuzza. Introduction to autonomous mobile robots. MIT press, (2011).

[19] Thrun, Sebastian, Wolfram Burgard, and Dieter Fox. *Probabilistic robotics*. MIT press, 2005.

[20] Crowley, James L. "World modeling and position estimation for a mobile robot using ultrasonic ranging." Robotics and Automation, 1989. Proceedings., 1989 IEEE International Conference on. IEEE, 1989.

[21] Ayache, Nicholas, and Olivier D. Faugeras. "Building, registrating, and fusing noisy visual maps." The International Journal of Robotics Research 7.6 (1988): 45-65.

[22] Jordan, M., "An Introduction to Probabilistic Graphical Models", University of California, Berkeley, 2003.

[23] Cheeseman, P., R. Smith, and M. Self. "A stochastic map for uncertain spatial relationships." 4th International Symposium on Robotic Research. 1987.

[24] Bar-Shalom, Yaakov. *Tracking and data association*. Academic Press Professional, Inc., 1987.

[25] Neira, José, and Juan D. Tardós. "Data association in stochastic mapping using the joint compatibility test." IEEE Transactions on robotics and automation 17.6 (2001): 890-897.

[26] Stuart Russell and Peter Norvig, "Artificial Intelligence: A Modern Approach," Prentice-Hall, Englewood Cliffs, NJ, 2010,3rd edition.